\documentclass[10pt]{article} %
\usepackage[preprint]{tmlr}

\usepackage{hyperref}
\usepackage{url}

\usepackage{amsmath}
\usepackage{amssymb}
\usepackage{mathtools}
\usepackage{amsthm}

\usepackage[capitalize,noabbrev]{cleveref}

\usepackage{placeins}
\usepackage{caption}

\usepackage{xspace}

\usepackage{amsfonts} %
\usepackage{thmtools,thm-restate}
\usepackage{mathcommand}
\usepackage{nicefrac}       %
\usepackage{etoolbox}
\usepackage[bb=dsserif]{mathalpha}

\usepackage{subcaption}

\usepackage{pifont}
\usepackage{graphicx}
\usepackage{booktabs} %
\usepackage{fancyhdr}
\usepackage{multirow}

\usepackage{tikz}
\usepackage{tcolorbox}
\tcbuselibrary{breakable}

\hypersetup{
    colorlinks=true,
    linkcolor=blue,
    citecolor=blue,
    urlcolor=blue,
    linktoc=all,
    pdfstartview=FitH,
    bookmarksopen=true,
    bookmarksopenlevel=2,
    bookmarksnumbered=true,
    bookmarksopen=true,
    bookmarksopenlevel=2,
    bookmarksnumbered=true,
    pdfpagemode=UseOutlines,
    pdfstartview=FitH,
    pdfpagelayout=OneColumn,
    pdfborder={0 0 0},
    pdfnewwindow=true,
    pdfauthor={Andreas Kirsch},
}

\usepackage{comment}
\excludecomment{draft}
\excludecomment{nextversion}

\theoremstyle{plain}

\theoremstyle{definition}

\theoremstyle{remark}

\providecommand\given{\MidSymbol[\vert]}

\newcommand\MidSymbol[1][]{%
\nonscript\:#1
\allowbreak
\nonscript\:
\mathopen{}}

\DeclareMathOperator{\opVar}{Var}

\DeclarePairedDelimiterXPP{\Var}[2]{\opVar_{#1}}{[}{]}{}{%
    \renewcommand\given{\MidSymbol[\delimsize\vert]}
    \ifblank{#2}{\:\cdot\:}{#2}
}

\DeclareMathOperator{\opExpectation}{\mathbb{E}}

\DeclarePairedDelimiterXPP{\implicitE}[1]{\opExpectation}{[}{]}{}{%
    \renewcommand\given{\MidSymbol[\delimsize\vert]}
    \ifblank{#1}{\:\cdot\:}{#1}
}

\DeclarePairedDelimiterXPP{\E}[2]{\opExpectation_{#1}}{[}{]}{}{%
    \renewcommand\given{\MidSymbol[\delimsize\vert]}
    \ifblank{#2}{\:\cdot\:}{#2}
}

\DeclareMathOperator{\opCovariance}{\mathrm{Cov}}

\DeclarePairedDelimiterXPP{\implicitCov}[1]{\opCovariance}{[}{]}{}{%
    \renewcommand\given{\MidSymbol[\delimsize\vert]}
    \ifblank{#1}{\:\cdot\:}{#1}
}

\DeclarePairedDelimiterXPP{\Cov}[2]{\opCovariance_{#1}}{[}{]}{}{%
    \renewcommand\given{\MidSymbol[\delimsize\vert]}
    \ifblank{#2}{\:\cdot\:}{#2}
}

\DeclarePairedDelimiterXPP{\indicator}[1]{\mathbb{1}}{\{}{\}}{}{%
    \ifblank{#1}{\:\cdot\:}{#1}
}

\DeclareMathOperator{\opInformationContent}{H}
\DeclarePairedDelimiterXPP{\ICof}[1]{\opInformationContent}{(}{)}{}{%
    \ifblank{#1}{\:\cdot\:}{#1}
}

\DeclareMathOperator{\opEntropy}{H}
\DeclarePairedDelimiterXPP{\Hof}[1]{\opEntropy}{[}{]}{}{%
    \renewcommand\given{\MidSymbol[\delimsize\vert]}
    \ifblank{#1}{\:\cdot\:}{#1}
}
\DeclarePairedDelimiterXPP{\xHof}[1]{\opEntropy}{(}{)}{}{%
    \ifblank{#1}{\:\cdot\:}{#1}
}

\DeclareMathOperator{\opMI}{I}
\DeclarePairedDelimiterXPP{\MIof}[1]{\opMI}{[}{]}{}{%
    \renewcommand\given{\MidSymbol[\delimsize\vert]}
    \ifblank{#1}{\:\cdot\:}{#1}
}

\DeclareMathOperator{\opTC}{TC}
\DeclarePairedDelimiterXPP{\TCof}[1]{\opTC}{[}{]}{}{%
    \renewcommand\given{\MidSymbol[\delimsize\vert]}
    \ifblank{#1}{\:\cdot\:}{#1}
}

\DeclarePairedDelimiterXPP{\CrossEntropy}[2]{\opEntropy}{(}{)}{}{%
    \ifblank{#1#2}{\:\cdot\: \MidSymbol[\delimsize\Vert] \:\cdot\:}{#1 \MidSymbol[\delimsize\Vert] #2}
}

\DeclareMathOperator{\opKale}{D_\mathrm{KL}}
\DeclarePairedDelimiterXPP{\Kale}[2]{\opKale}{(}{)}{}{%
    \ifblank{#1#2}{\:\cdot\: \MidSymbol[\delimsize\Vert] \:\cdot\:}{#1 \MidSymbol[\delimsize\Vert] #2}
}

\DeclareMathOperator{\opp}{p}
\DeclarePairedDelimiterXPP{\pof}[1]{\opp}{(}{)}{}{%
    \renewcommand\given{\MidSymbol[\delimsize\vert]}
    \ifblank{#1}{\:\cdot\:}{#1}
}

\DeclarePairedDelimiterXPP{\pcof}[2]{\opp_{#1}}{(}{)}{}{%
    \renewcommand\given{\MidSymbol[\delimsize\vert]}
    \ifblank{#2}{\:\cdot\:}{#2}
}

\DeclarePairedDelimiterXPP{\pfor}[1]{\opp}{\lbrack}{\rbrack}{}{%
    \renewcommand\given{\MidSymbol[\delimsize\vert]}
    \ifblank{#1}{\:\cdot\:}{#1}
}

\DeclarePairedDelimiterXPP{\pcfor}[2]{\opp_{#1}}{\lbrack}{\rbrack}{}{%
    \renewcommand\given{\MidSymbol[\delimsize\vert]}
    \ifblank{#2}{\:\cdot\:}{#2}
}

\DeclarePairedDelimiterXPP{\hpcof}[2]{\hat{\opp}_{#1}}{(}{)}{}{%
    \renewcommand\given{\MidSymbol[\delimsize\vert]}
    \ifblank{#2}{\:\cdot\:}{#2}
}

\DeclareMathOperator{\opq}{q}
\DeclarePairedDelimiterXPP{\qof}[1]{\opq}{(}{)}{}{%
    \renewcommand\given{\MidSymbol[\delimsize\vert]}
    \ifblank{#1}{\:\cdot\:}{#1}
}

\DeclarePairedDelimiterXPP{\qcof}[2]{\opq_{#1}}{(}{)}{}{%
    \renewcommand\given{\MidSymbol[\delimsize\vert]}
    \ifblank{#2}{\:\cdot\:}{#2}
}

\DeclarePairedDelimiterXPP{\varHof}[2]{\opEntropy_{\ifblank{#1}{\:\cdot\:}{#1}}}{[}{]}{}{%
    \renewcommand\given{\MidSymbol[\delimsize\vert]}
    \ifblank{#2}{\:\cdot\:}{#2}
}

\DeclarePairedDelimiterXPP{\xvarHof}[2]{\opEntropy_{\ifblank{#1}{\:\cdot\:}{#1}}}{(}{)}{}{%
    \renewcommand\given{\MidSymbol[\delimsize\vert]}
    \ifblank{#2}{\:\cdot\:}{#2}
}

\DeclarePairedDelimiterXPP{\aICof}[1]{\opInformationContent'}{(}{)}{}{%
    \ifblank{#1}{\:\cdot\:}{#1}
}

\DeclarePairedDelimiterXPP{\aHof}[1]{\opEntropy'}{[}{]}{}{%
    \renewcommand\given{\MidSymbol[\delimsize\vert]}
    \ifblank{#1}{\:\cdot\:}{#1}
}
\DeclarePairedDelimiterXPP{\axHof}[1]{\opEntropy'}{(}{)}{}{%
    \ifblank{#1}{\:\cdot\:}{#1}
}

\DeclarePairedDelimiterXPP{\aMIof}[1]{\opMI'}{[}{]}{}{%
    \renewcommand\given{\MidSymbol[\delimsize\vert]}
    \ifblank{#1}{\:\cdot\:}{#1}
}

\DeclarePairedDelimiterXPP{\aCrossEntropy}[2]{\opEntropy'}{(}{)}{}{%
    \ifblank{#1#2}{\:\cdot\: \MidSymbol[\delimsize\Vert] \:\cdot\:}{#1 \MidSymbol[\delimsize\Vert] #2}
}

\DeclarePairedDelimiterXPP{\HofHessian}[1]{\opEntropy''}{[}{]}{}{%
    \renewcommand\given{\MidSymbol[\delimsize\vert]}
    \ifblank{#1}{\:\cdot\:}{#1}
}

\DeclarePairedDelimiterXPP{\specialHofHessian}[2]{\opEntropy''#1}{[}{]}{}{%
    \renewcommand\given{\MidSymbol[\delimsize\vert]}
    \ifblank{#2}{\:\cdot\:}{#2}
}

\DeclarePairedDelimiterXPP{\HofJacobian}[1]{\opEntropy'}{[}{]}{}{%
    \renewcommand\given{\MidSymbol[\delimsize\vert]}
    \ifblank{#1}{\:\cdot\:}{#1}
}

\DeclarePairedDelimiterXPP{\specialHofJacobian}[2]{\opEntropy'#1}{[}{]}{}{%
    \renewcommand\given{\MidSymbol[\delimsize\vert]}
    \ifblank{#2}{\:\cdot\:}{#2}
}

\newcommand{\Yacq}{{Y^\text{acq}}}

\newcommand{\x}{{x}}
\newcommand{\Y}{{Y}}
\newcommand{\y}{{y}}

\newcommand{\W}{\Omega}

\newcommand{\andreas}[1]{}
\newcommand{\yarin}[1]{}
\newcommand{\forreviewers}[1]{}

\begin{draft}
\renewcommand{\andreas}[1]{{\leavevmode\color{blue}{ \footnotesize AK} {\tiny says: }#1}}
\renewcommand{\yarin}[1]{{\leavevmode\color{orange}{ \footnotesize YG} {\tiny says: }#1}}
\renewcommand{\forreviewers}[1]{{\leavevmode\color{purple}{\footnotesize NOTE} {\tiny please: }#1}}
\end{draft}

\newmathcommand{\batchvar}{B}
\newtextcommand{\batchvar}{$\batchvar$\xspace}

\newmathcommand{\poolsize}{P}
\newtextcommand{\poolsize}{$\poolsize$\xspace}

\newmathcommand{\trainsize}{N}
\newtextcommand{\trainsize}{$\trainsize$\xspace}

\newmathcommand{\evalsize}{E}
\newtextcommand{\evalsize}{$\evalsize$\xspace}

\newmathcommand{\testsize}{M}
\newtextcommand{\testsize}{$\testsize$\xspace}

\newmathcommand{\numclasses}{K}
\newtextcommand{\numclasses}{$\numclasses$\xspace}

\everypar{\looseness=-1}
\allowdisplaybreaks

\newtcolorbox{importantresult}{colback=solarized@yellow!5!white,
colframe=solarized@yellow,parbox, left=0.5mm, right=0.5mm,top=0.5mm,bottom=0.5mm}

\newtcolorbox{mainresult}{colback=solarized@violet!5!white,
colframe=solarized@violet,parbox, left=0.5mm, right=0.5mm,top=0.5mm,bottom=0.5mm}

\newtcolorbox{importantresult_noparbox}{breakable,colback=solarized@yellow!5!white,
colframe=solarized@yellow,parbox=false, left=0.5mm, right=0.5mm,top=0.5mm,bottom=0.5mm}

\title{Speeding Up BatchBALD: A k-BALD Family of Approximations for Active Learning}

\author{\name Andreas Kirsch \email andreas.kirsch@exeter.ox.ac.uk \\
      \addr OATML, Department of Computer Science\\
      University of Oxford
}

\def\month{MM}  %
\def\year{YYYY} %
\def\openreview{\url{https://openreview.net/forum?id=XXXX}} %

\begin{document}

\maketitle

\begin{abstract}
  Active learning is a powerful method for training machine learning models with limited labeled data.
  One commonly used technique for active learning is BatchBALD, which uses Bayesian neural networks to find the most informative points to label in a pool set. 
  However, BatchBALD can be very slow to compute, especially for larger datasets.
  In this paper, we propose a new approximation, k-BALD, which uses k-wise mutual information terms to approximate BatchBALD, making it much less expensive to compute.
  Results on the MNIST dataset show that k-BALD is significantly faster than BatchBALD while maintaining similar performance.
  Additionally, we also propose a dynamic approach for choosing k based on the quality of the approximation, making it more efficient for larger datasets.
\end{abstract}

\section{Introduction}
Our paper addresses the issue of slow computation time for BatchBALD, a method for active learning that finds the most informative points to label in a pool set using Bayesian neural networks. We propose a new approximation, k-BALD, which uses k-wise mutual information terms to approximate BatchBALD, making it much less expensive to compute. As future work, we propose that the acquisition batch size or the order of approximation could be dynamically chosen based on its quality, making it more efficient for larger datasets.

The goal of active learning \citep{cohn1994improving} is to identify the most informative points in an unlabeled pool set to be labeled and added to the training set, in order to improve the performance of the model. One commonly used technique for active learning is BatchBALD \citep{kirsch2019batchbald}, which uses Bayesian neural networks to find the most informative points to label in a pool set. However, BatchBALD can be very slow to compute, especially for larger datasets. In this paper, we propose a new approximation, k-BALD, which uses k-wise mutual information terms to approximate BatchBALD, making it much less expensive to compute.

From an information-theoretic point of view, active learning means finding the unlabeled points in the pool set with the highest expected information gain, which is also referred to as BALD score \citep{houlsby2011bayesian}. BALD score is often used to capture the epistemic uncertainty of the model for a given point. When using Bayesian neural networks, BALD scores measure the disagreement between (Monte-Carlo) parameter samples, similar to `Query by Committee' \citep{seung1992query}. To be more specific, the BALD scores look as follows, where $q(\omega)$ is an empirical distribution of the parameters of the ensemble members, or an approximate parameter distribution, e.g., using Monte-Carlo dropout \citep{gal2015dropout}:
\begin{align}
\MIof{\W; \Y \given \x} &= \Hof{\Y \given \x} - \Hof{\Y \given \x, \W} = \int q(\omega) log \frac{p(\y \given \x, \omega)}{\int p(\y \given \x, \omega)q(\omega)d\y} d\omega.
\end{align}

BatchBALD, proposed in \citet{kirsch2019batchbald}, is an extension of the BALD algorithm to handle batch acquisition of multiple points at once. In practice, BatchBALD computes the \emph{joint} BALD score and uses the greedy algorithm from submodular optimization theory to build an acquisition batch that is $1-\nicefrac{1}{e}$-optimal:
\begin{align}
    \MIof{\W; \Yacq_1, \ldots \Yacq_k \given \x} = \Hof{\Yacq_1, \ldots \Yacq_k \given \x} - \Hof{\Yacq_1, \ldots \Yacq_k \given \x, \W}.
\end{align}
Labels for the points in the acquisition batch are then queried and added to the training set. 
In practice, however, computing BatchBALD can be very slow, especially for large datasets. 

In this paper\footnote{This research idea and initial results were published as a blog post initially: \url{https://web.archive.org/web/20220702232856/https://blog.blackhc.net/2022/07/kbald/}.}, we propose a new family of approximations for BatchBALD, called \emph{k-BALD}, which uses up to k-wise mutual information terms, leading to a much less expensive approximation.
For example, on MNIST \citep{deng2012mnist}, 2-BALD takes 1 min to select an acquisition batch of size 5, and at acquisition batch size 10, 2-BALD takes 2 min while it still performs as well as BatchBALD, while BatchBALD takes 1 min for acquisition batch size 5 and already 30 min for acquisition batch size 10, see also \Cref{fig:mnist_kbald_vs_batchbald}.
Importantly, we could use this family of approximations to dynamically choose the acquisition batch size by estimating the quality of our approximation---a first in active learning.

The rest of the paper is organized as follows. In section \S\ref{sec:method}, we will describe the k-BALD method, including the explanation of k-wise mutual information terms, how k-BALD approximates BatchBALD using these terms, and the dynamic choice of k based on approximation quality.
In section \S\ref{sec:results}, we will present our experiment results, including the comparison of computation time between BatchBALD and k-BALD on MNIST dataset and the comparison of performance between BatchBALD and k-BALD. Finally, we will conclude in section \S\ref{sec:conclusion}, summarizing our results and discussing future work and potential extensions.

\section{k-BALD: Application of the Inclusion-Exclusion Principle}
\label{sec:method}

In this section, we describe the k-BALD method for approximating BatchBALD. Our approach is based on the inclusion-exclusion principle, which allows us to approximate the BatchBALD score using k-wise mutual information terms. 
That is instead of trying to compute the joint entropy in the BatchBALD score exactly, we can approximate it using pairwise mutual information terms, leading to a new approximation, we call 2-BALD, or generally, following the inclusion-exclusion principle, using up to k-wise mutual information terms, leading to what call the k-BALD family of approximations for BatchBALD.
Additionally, we also propose a dynamic approach for choosing k based on the quality of the approximation.

\textbf{Inclusion-Exclusion Principle.} From set theory, it is known that for sets $S_i$, we have:
\begin{align}
    | \bigcup_i S_i | = \sum_i | S_i | - \sum_{i<j} | S_i \cap S_j | + \sum_{i<j<k} | S_i \cap S_j \cap S_k | + \cdots
\end{align}

Following \citet{yeung1991new}, which connects set operations with information quantities\footnote{see also \href{https://www.blackhc.net/blog/2019/better-intuition-for-information-theory/}{`Better intuition for information theory'} by your truly for an easy-going introduction.}, we can apply the same principle to information-theoretic quantities, which leads to the following decomposition:
\begin{align}    
    H[Y_1, \ldots, Y_B \mid x_1, \ldots x_B] = \sum_i H[Y_i \mid x_i] - \sum_{i < j} I[Y_i ; Y_j \mid x_i, x_j] + \sum_{i<j<k} I[Y_i ; Y_j ; Y_k \mid x_i, x_j, x_k] - \cdots
\end{align}
and therefore:
\begin{align}
    &I[Y_1, \ldots, Y_B ; \Omega \mid x_1, \ldots x_B] \notag \\
    &\quad = H[Y_1, \ldots, Y_B \mid x_1, \ldots x_B] - \sum_i H[Y_i \mid x_i, \Omega] \\
    &\quad = \sum_i I[Y_i ; \Omega \mid x_i] - \sum_{i < j} I[Y_i ; Y_j \mid x_i, x_j] + \sum_{i<j<k} I[Y_i ; Y_j ; Y_k \mid x_i, x_j, x_k] - \cdots    
\end{align}

In particular, we introduce the following approximations to BatchBALD:
\begin{itemize}
    \item \textbf{1-BALD}: $\sum_i I[Y_i; \Omega \mid x_i ]$
    \item \textbf{2-BALD}: $\sum_i I[Y_i; \Omega \mid x_i ] - \sum_{i < j} I[Y_i ; Y_j \mid x_i, x_j]$
    \item \textbf{...}
    \item \textbf{k-BALD}: $\sum_i I[Y_i ; \Omega \mid x_i] - \sum_{i < j} I[Y_i ; Y_j \mid x_i, x_j] + \sum_{i<j<k} I[Y_i ; Y_j ; Y_k \mid x_i, x_j, x_k] - \cdots$
\end{itemize}

With 1-BALD, we simply recover the well-known top-K BALD, where we greedily maximize over the possible candidates for the acquisition batch using individual BALD scores.

\textbf{Total Correlation.} The following relationship between BatchBALD and the 1-BALD scores is straightforward to derive:
$$I[\Omega; Y_1, \ldots, Y_B \mid x_1, \ldots, x_B] =  \sum_i I[\Omega; Y_i \mid x_i] - TC[Y_1 ; \ldots ; Y_B \mid x_1 ; \ldots ; x_B],$$
where the total correlation is:
$$TC[Y_1 ; \ldots ; Y_B \mid x_1 ; \ldots ; x_B] = \sum_i H[Y_i \mid x_i] - H[Y_1, \ldots, Y_B \mid x_1, \ldots, x_B].$$
The total correlation measures the dependence of the predictions on each other. When it is $0$, the random variables are independent.

While BatchBALD fully estimates this total correlation, k-BALD for $k<K$ approximates the total correlation using the k-wise mutual information terms:
$$
TC[Y_1 ; \ldots ; Y_B \mid x_1 ; \ldots ; x_B] \approx \sum_{i < j} I[Y_i ; Y_j \mid x_i, x_j] - \sum_{i<j<k} I[Y_i ; Y_j ; Y_k \mid x_i, x_j, x_k] + \cdots.
$$
So while 1-BALD does not take the total correlation into account at all; 2-BALD takes the total correlation into account up to pairwise terms; and k-BALD takes the total correlation into account up to k-wise terms.

Indeed, from statistical learning theory, we know that in the infinite training data limit, the model parameters converge, and the predictions become independent of each other.
As the total correlation decreases, 1-BALD becomes closer and closer to BatchBALD. 
We can conjecture that 1-BALD is a good approximation of BatchBALD in this limit, but not early in the active learning process.

\textbf{Dynamic Acquisition Batch Sizes.} An important advantage of k-BALD is that it allows us to dynamically choose the acquisition batch size by estimating the quality of our approximation. 
We can hope that later in training, we could increase the batch size automatically without loss in label efficiency. Crucially, this depends on the total correlation decreasing further along in training.

For example, we could compute both 2-BALD and 3-BALD and stop the batch acquisition once the scores of 2- and 3-BALD diverge too much. 

\section{Evaluation Results}
\label{sec:results}

In this section, we present the results of our experiments comparing the performance of BatchBALD and 2-BALD on the MNIST dataset. We also discuss the challenges and limitations that arise from our initial results. We follow the codebase from \citet{kirsch2019batchbald} and use the same experimental setup for MNIST and the same hyperparameters.

\subsection{Comparison of BatchBALD and 2-BALD}

\begin{figure}[t]
\centering
\begin{minipage}[t]{0.49\textwidth}
\centering
\includegraphics[width=\textwidth]{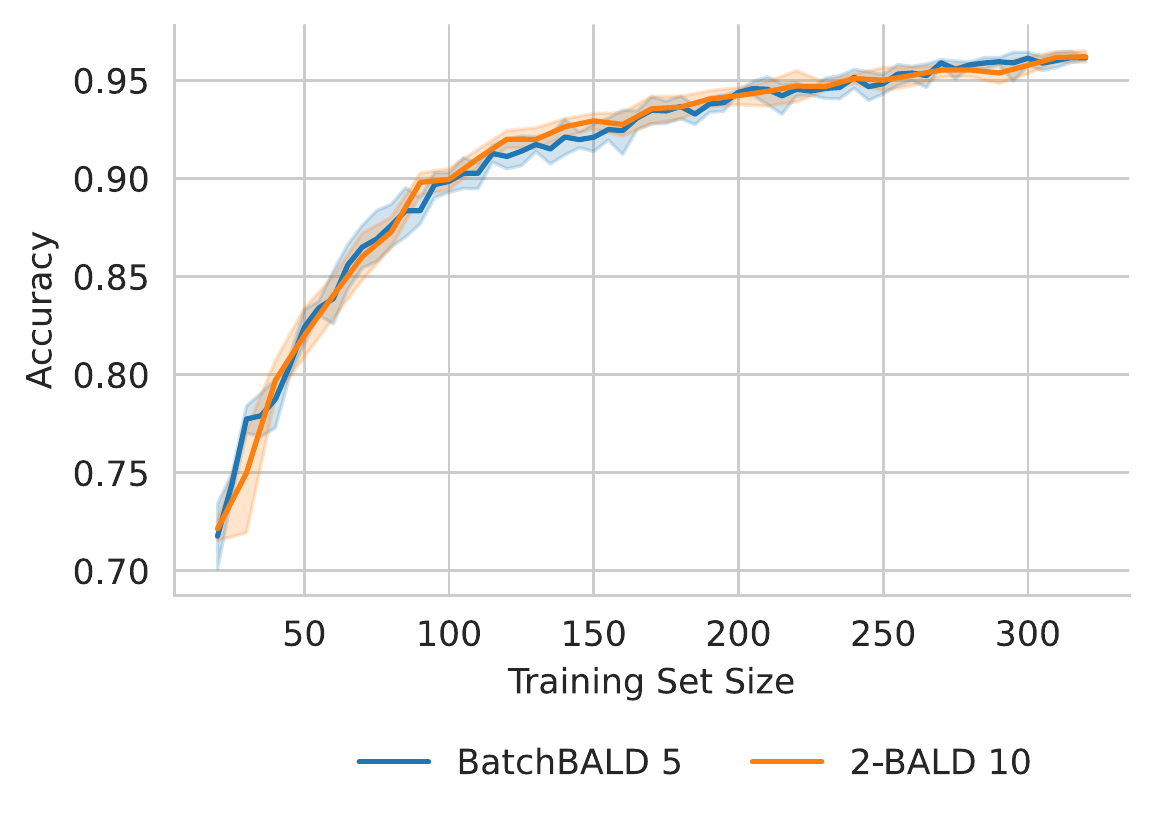}
\caption{Performance of 2-BALD on MNIST. 2-BALD performs as well as BatchBALD for a fraction of the computational cost.}
\label{fig:mnist_kbald_vs_batchbald}
\end{minipage}
\begin{minipage}[t]{0.49\textwidth}
\centering
\includegraphics[width=\textwidth]{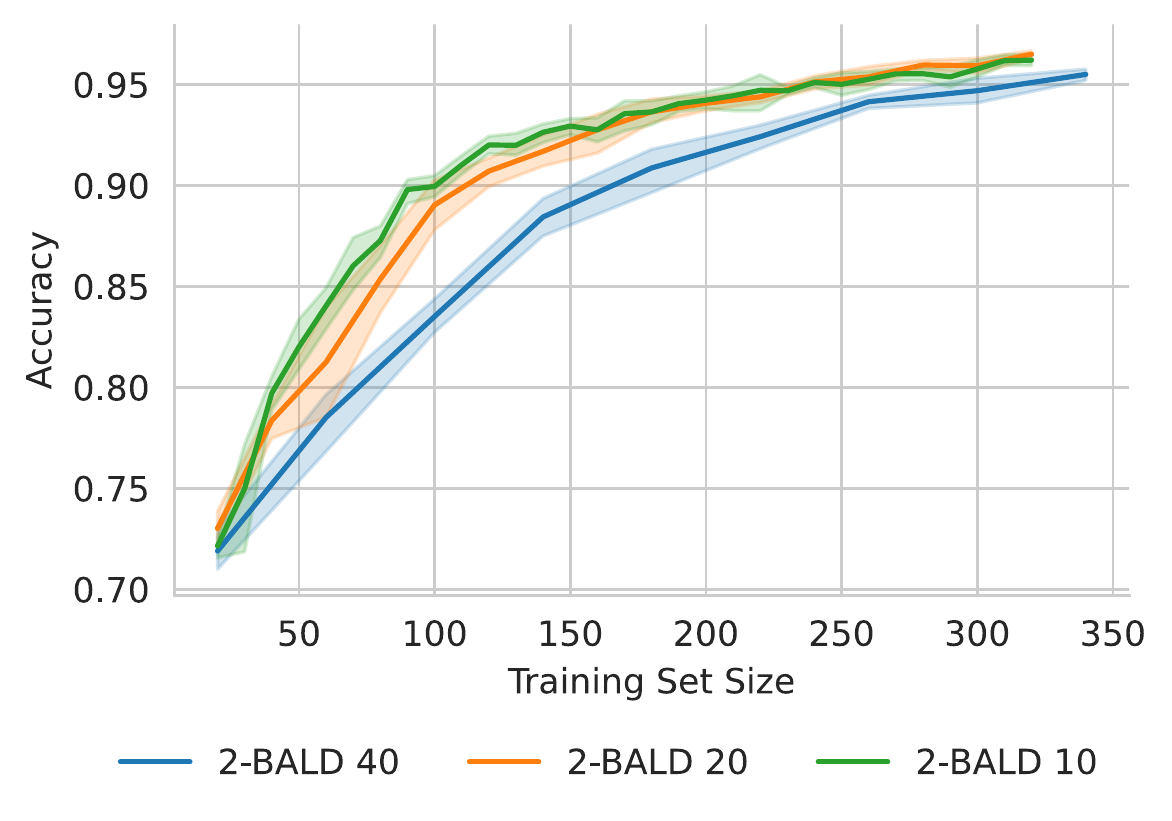}
\caption{Effect of the acquisition batch size on the performance of 2-BALD. 2-BALD's performance per acquisition round does not improve with larger acquisition batch sizes.}
\label{fig:mnist_acquisition_size_ablation}
\end{minipage}
\end{figure}

\begin{figure*}[t]
    \centering
    \begin{minipage}[t]{0.60\textwidth}
        \centering
        \includegraphics[width=\textwidth]{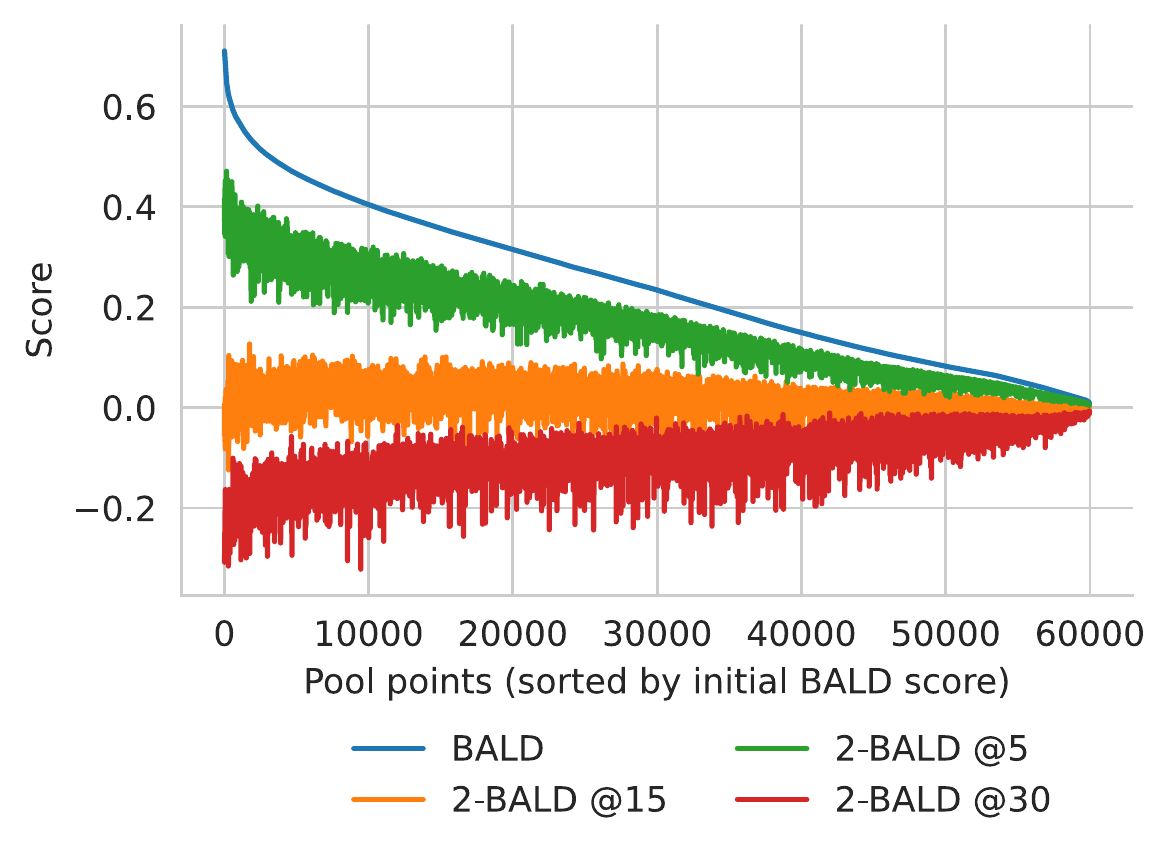}
        \caption{Progression of 2-BALD scores. Subtracting interactions between pairs of samples eventually makes the originally most informative points the most negative. Thus degrading to worse performance than random acquisition (not shown here).}
        \label{fig:mnist_kbald_score_progression}
    \end{minipage} \hfill
    \begin{minipage}[b]{0.39\textwidth}
        \centering
        \begin{tabular}{lcc}
        \toprule
        \multirow{2}{*}{\textbf{Method}} & \multicolumn{2}{c}{\textbf{Acq. Time (min)}} \\
        \cmidrule(lr){2-3}
        & \textbf{Acq. Size 5} & \textbf{10} \\
        \midrule
        BatchBALD & 1 & 30 \\
        2-BALD & 1 & 2 \\
        \bottomrule
        \end{tabular}
        \captionof{table}{Time taken to select an acquisition batch of size 5 and 10 for BatchBALD and 2-BALD, on an Nvidia Geforce 2080 Super.}
        \label{tab:mnist_time}
    \end{minipage} %
\end{figure*}

We first evaluate the performance of 2-BALD using an acquisition batch size of 5 for BatchBALD and 10 for 2-BALD. The results are shown in \Cref{fig:mnist_kbald_vs_batchbald}. As can be seen from the figure, 2-BALD performs as well as BatchBALD in terms of both accuracy and performs much better in regards to computation time:
In \Cref{tab:mnist_time}, 2-BALD takes 1 min to select an acquisition batch of size 5, and at acquisition batch size 10, 2-BALD takes 2 min while still performs as well as BatchBALD, at least in the proof of concept experiment on MNIST, while BatchBALD takes 1 min for acquisition batch size 5 and already 30 min for acquisition batch size 10.

\FloatBarrier
    
\subsection{Limitations of 2-BALD with Larger Acquisition Batch Sizes}

However, when we increase the acquisition batch size, the performance of 2-BALD deteriorates. \Cref{fig:mnist_acquisition_size_ablation} shows the results of an ablation study on the effect of acquisition batch size on 2-BALD's performance. As can be seen, 2-BALD performs poorly for larger acquisition batch sizes.

We have examined why this is the case and found that as we add additional samples and subtract the pairwise interactions, we subtract too much, even pushing the scores to become negative. This can be seen in \Cref{fig:mnist_kbald_score_progression}, where the 2-BALD scores eventually become negative. As a result, 2-BALD prefers uninformative points later in the acquisition batch.

This represents a limitation for 2-BALD as it will eventually acquire uninformative points, leading to a degradation of performance compared to even random acquisition. This means that while 2-BALD might be a viable alternative to BatchBALD when using comparable acquisition batch sizes, it does not allow for scalability to larger acquisition batch sizes. However, on larger datasets with more classes, it is possible that larger acquisition sizes would be viable.

\section{Conclusion}
\label{sec:conclusion}

In this paper, we introduced a new family of approximations for BatchBALD, k-BALD, that use k-wise mutual information terms to approximate BatchBALD. k-BALD is much less expensive to compute than BatchBALD, and can be dynamically chosen based on the quality of the approximation. Results on the MNIST dataset showed that k-BALD is significantly faster than BatchBALD while maintaining similar performance.

\textbf{Future Work.} %
There are several interesting avenues for this idea. One possible direction is to investigate the relationship between k-BALD and BatchBALD as the order of the k-wise mutual information terms increases. This could provide insight into the performance of k-BALD as the approximation approaches the full joint mutual information. Additionally, it may be worthwhile to investigate the divergence of 3-BALD from 2-BALD as a way to dynamically set the acquisition batch size and catch approximation issues.

Another interesting avenue for future work is to investigate the use of k-BALD for ``conservative'' acquisition batches. This approach involves greedily acquiring the highest BALD scorer and removing all pool samples from future consideration in that acquisition round which have total correlation greater than a threshold value, repeating the process until no acceptable samples remain. This is a more conservative approach, but may be useful for situations where it is important to make only ``conservative'' acquisitions. Importantly, we could also use this approach to dynamically set the acquisition batch size. For example, we could set the threshold value to be the maximum total correlation between any two samples in the acquisition batch and use that as overall ``bugdet''. This would allow us to dynamically set the acquisition batch size, too.

Finally, it is also worth considering the quality of the posterior approximation when using k-BALD for active learning. In order to predict many points into the future, it is necessary to have a good posterior approximation and to sample diverse predictions from the posterior. This may present a challenge for the scalability of k-BALD and is an important area for future research.

Overall, the k-BALD family of approximations for BatchBALD presents an exciting new direction for active learning, providing a more efficient and scalable alternative to traditional BatchBALD methods.

\textbf{Concurrent Work.} We would like to acknowledge the recent work of Rubashevskii, Kotova, and Panov in their paper ``Scalable Batch Acquisition for Deep Bayesian Active Learning'' \citep{rubashevskii2023scalable}. They propose a similar method, Large BatchBALD, which also uses 2-wise mutual information terms to approximate BatchBALD and improve computational efficiency. However, their approach combines Large BatchBALD with stochastic batch acquisitions \citep{kirsch2021stochastic} and shows that it performs similarly to or better than PowerBALD from \citep{kirsch2021stochastic}.

It is important to note that while their work and our proposed k-BALD method share similar ideas and insights, there are key differences in the scope and focus of our respective papers. Specifically, they do not examine how the scores change within an acquisition round and the quality of the approximation  (see \Cref{fig:mnist_kbald_score_progression}), and we do not consider the dynamic setting of the acquisition batch size or conservative acquisition.

It is worth noting that our research idea and the Large BatchBALD method were developed concurrently\footnote{This research idea and initial results were published as a blog post initially in July 2022, see \url{https://web.archive.org/web/20220702232856/https://blog.blackhc.net/2022/07/kbald/}.}, and it was brought to our attention that the authors of the Scalable Batch Acquisition paper were not aware of our work when they submitted their paper (and same vice-versa). They have stated that they will update their arXiv version to include reference to our work. Our goal in publishing this note on our research idea is to make it more citable and to contribute to the ongoing conversation on efficient methods for active learning.

\section*{Acknowledgements}

AK is supported by the UK EPSRC CDT in Autonomous Intelligent Machines and Systems (grant reference EP/L015897/1).

\bibliography{main}
\bibliographystyle{tmlr}

\end{document}